\UseRawInputEncoding
\documentclass[journal=apchd5, manuscript=article]{achemso}
\usepackage[utf8]{inputenc}
\usepackage{upquote}
\usepackage{newunicodechar}
\newunicodechar{，}{,}
\newunicodechar{•}{\textbullet{}}
\newunicodechar{≤}{\leq{}}
\newunicodechar{Δ}{\ensuremath{\Delta}}
\usepackage{amssymb}
\usepackage{algorithm}
\usepackage{algorithmicx}
\usepackage{algpseudocode}
\usepackage{amssymb}
\usepackage[most]{tcolorbox}
\usepackage[utf8]{inputenc}    
\usepackage{listings}          
\usepackage{array} 
\usepackage{booktabs}
\usepackage[version=3]{mhchem} 

\author{Darui Lu}
\affiliation[Duke University]
{Department of Electrical and Computer Engineering, Duke University, Durham, USA}

\author{Jordan M. Malof}
\affiliation[Duke University]
{Department of Electrical Engineering and Computer Science, University of Missouri, Columbia, USA}

\author{Willie J. Padilla}
\email{willie.padilla@duke.edu}
\affiliation[Duke University]
{Department of Electrical and Computer Engineering, Duke University, Durham, USA}

\title[An \textsf{achemso} demo]
  {An Agentic Framework for Autonomous Metamaterial Modeling and Inverse Design}

\abbreviations{IR,NMR,UV}
\keywords{Artificial intelligence, photonics, metamaterials, metasurfaces, agents, inverse design}

\begin{document}

\begin{tocentry}

  \includegraphics[scale=0.85]{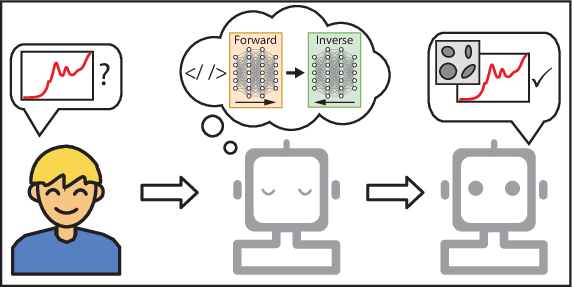} 

\end{tocentry}

\begin{abstract}
  The evolution from large language models (LLMs) to agentic systems has created a new frontier of scientific discovery, enabling the automation of complex research tasks that have traditionally required human expertise. We develop and demonstrate such a framework specifically for the inverse design of photonic metamaterials. When queried with a desired optical spectrum, the \textit{Agent} autonomously proposes and develops a forward deep learning model, accesses external tools via APIs for tasks like optimization, utilizes memory, and generates a final design \textit{via} a deep inverse method. We demonstrate the framework's effectiveness, highlighting its ability to reason, plan, and adapt its strategy autonomously and in real-time, mirroring the processes of a human researcher. Notably, the \textit{Agentic Framework} possesses internal reflection and decision flexibility, permitting highly varied and potentially novel outputs.
\end{abstract}

\section{Introduction}

Deep learning models, such as Deep Neural Networks (DNNs), have shown great promise in photonics research\cite{Khatib2021,Qian2025,Cerniauskas2024} including metamaterials\cite{zhang2024}, metasurfaces\cite{Zhang2023}, plasmonics\cite{Malkiel2018}, and photonic crystals\cite{Ma2020}.  One notable application is that of \textit{forward modeling} where DNNs have been found to accurately predict the properties of metamaterials based upon their design (e.g., geometric properties).  Many prior works employ fully connected networks \cite{deng2021neural,hou2020prediction,peurifoy2018nanophotonic} and convolutional neural networks are also a popular choice, particularly for free-form metamaterials\cite {kiymik2022metamaterial,jiang2022dispersion,zhang2021deep}. Other types of networks have also been explored, including transformers\cite{chen2023broadband,vit_meta,deng2021benchmarking} and graphical neural networks\cite{Khoram2022}. Another notable application is \textit{inverse design}, where a DNN is trained to take the desired properties of a metamaterial as input and return the design needed to obtain those properties. A variety of DNN-based inverse design methods have been proposed such as diffusion models\cite{Zhang2023}, tandem\cite{Xu2023}, and neural adjoint (NA)\cite{deng2021neural}. We refer the reader to recent reviews (e.g., \cite{Ren2022}) or tutorials (e.g., \cite{KhairehWalieh2023, Deng2022t}) for a broader discussion of such methods.

For both forward and inverse modeling, a human researcher must make numerous design choices: creating an overall plan, choosing simulation tools and settings for dataset generation, choosing the appropriate machine learning algorithms and their hyperparameters.  Once these design choices are made, the resulting process must be implemented in specialized software.  The design of the algorithms is then typically adjusted iteratively in response to intermediate experimental results.  Although these machine learning processes enable the modeling and design of advanced materials, they also require substantial time, which slows scientific progress.  The expertise required to set up these procedures also limits the accessibility of these advanced pipelines.  These challenges are not unique to metamaterial research, but extend to scientific computing more broadly, where similar machine learning processes have experienced success (e.g., \cite{jumper2021highly,azizzadenesheli2024neural,walters2020applications}).

To solve these challenges, we propose to leverage large language model (LLM)-based agentic systems. An agentic framework could replace human decision-making steps in the aforementioned process, dramatically reducing the barrier to adopting advanced machine learning procedures, and relieving human scientists of the time-consuming aspects of the pipeline. In this work, we introduce and validate a novel Agentic Framework capable of performing inverse design of metamaterial photonic systems. The primary contributions of this work are as follows:

\begin{itemize}
    \item We introduce an Agentic Framework that automates the complete end-to-end process of metamaterial inverse design, a scope that includes the autonomous development and optimization of a DNN-based surrogate forward model.
    \item We demonstrate that a system of specialized LLM-based agents can collaborate to achieve similar performance to a human scientist in complex scientific tasks, evincing the power of an agentic approach.
    \item Our framework realizes a novel level of autonomy in scientific research by incorporating internal reflection and dynamic planning, allowing it to adapt its methodology based on intermediate results.
\end{itemize}

\section{The Agentic Framework}

We aim to develop an agentic system to automate an inverse design process that relies upon the Neural Adjoint (NA) inverse method, which is relatively easy to use and has achieved state-of-the-art performance on such modeling tasks \cite{Ren2022,ren2020benchmarking}. We first build an accurate DNN-based surrogate forward model, which can predict the properties of a metamaterial based upon its design. As discussed, this process of building a DNN-based forward model is now widely-adopted for advanced metamaterial research, and is therefore common practice in modern metamaterial laboratories. The forward modeling process typically consists of two alternating steps: data collection and DNN optimization. During data collection, pairs of designs and metamaterial properties are acquired via experimentation, which is either physical or (more often) numerical simulation. Optimization of the DNN utilizes this data to train and evaluate models. We will often consider multiple DNN model designs and select one that achieves the lowest error on the designated testing data.  These two steps of the forward modeling process are repeated -- not necessarily in equal proportion -- until a suitable level of prediction error is achieved. Once we have an accurate forward model, the NA design method prescribes how we can re-purpose it (i.e., without additional training) to perform inverse design, so that we can simply provide the model with some desired metamaterial properties and it returns a design that will achieve those properties. Here we present an autonomous agent that intelligently engages in the end-to-end metamaterial design process. Our agent replaces the need for manual intervention by dynamically planning and carrying out several complex tasks, which have historically required deep expertise in both photonics and deep learning.

To automate the targeted inverse design process, we investigate agentic systems that comprise a collection of large language models (LLMs).  LLMs are essentially DNNs that have been trained to take sequences of tokens as input - termed a \textit{prompt} - and return another sequence of tokens as output.  Each token is a numerical vector that can, in principle, represent various types of data such as text, audio, or imagery. In our context, however, a token will usually represent textual characters, and the input and output of the LLM will consist of textual data, such as natural language or and numerical tables. In recent years, LLMs have been shown capable of effectively addressing highly complex prompts, such as answering technical questions about the best machine learning algorithm for a given problem, to carrying out tasks like writing the code to achieve some outcome. In more recent research, LLMs have been prompted to use tools to solve tasks or pursue goals, as long as those tools can be manipulated via text (e.g., the LLM can output commands that run a simulator).

Despite these powerful emerging capabilities of LLMs, they remain limited in several ways. Although LLMs can exhibit some capability for reasoning, they often struggle with long-term planning of complex multi-step tasks that involve a long sequential set of textual input and output from the LLM. That is they tend to lose sight of the long-term plan while carrying out several tasks, and react more directly to their most recent input prompts. Furthermore LLMs posses a finite-sized context window explicitly limiting their ability to carry out long tasks. Although LLMs can be given access to tools, they might struggle to determine the correct sequence of tool-use and are poor at handling errors. Some state-of-the-art (SOTA) prompting strategies can offer a degree of mitigation for these limitations, and include zero / few-shot prompting, \cite{brown2020language} system prompting, role prompting, chain of thought, \cite{wei2022chain} self-consistency, \cite{wang2022self} and tree of thoughts \cite{yao2023tree}. However, despite the implementation of SOTA prompting techniques into LLMs, alternative \textit{Agentic} approaches offer much better performance for complex tasks. An Agentic Framework is a system of LLMs and tools working together to carry out complex tasks. By constructing a team of LLMs and giving them access to tools, they may together overcome many of the above listed limitations of LLMs. Although no universal definition exists, a consensus is beginning to emerging, suggesting that an Agentic Framework should be autonomous and goal-oriented, capable of reasoning, planning, and adapting through the use of tools and memory.

\begin{figure}[ptb]
    \centering
    \includegraphics[width=0.9\textwidth]{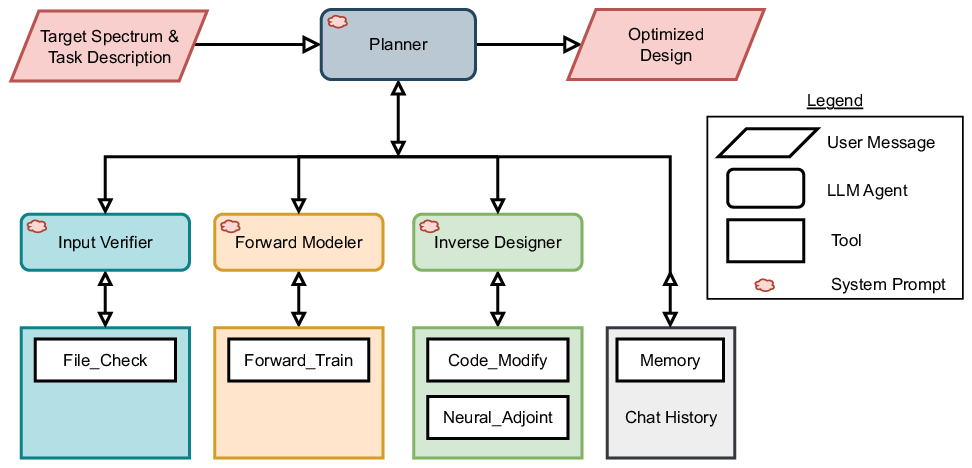}
    \caption{Schematic showing the Agentic Framework. The Planner is a general LLM which controls the entire Agentic process. The LLM Agent blocks (rounded rectangles) are specialized LLMs which are provided with system prompts (red thought bubbles) to enable specialized functions. Tools are APIs that allow the Agent to access features or data, perform simulations, and to store and recall short or long term memory (gray), and are termed: File$\_$Check, Forward$\_$Train, Code$\_$Modify, and Neural$\_$Adjoint.}
    \label{fig1}
\end{figure}

Our proposed Agentic Framework, depicted in Fig. \ref{fig1}, embodies these principles and possesses a dedicated planning module (Planner) that orchestrates the overall process including natural language processing (communicating with the user), LLM management, and tool use. Memory in our Agentic Framework is enabled with the OpenAI Agent SDK’s built-in memory store to record and retrieve chat history, permitting the Planner and other LLMs to maintain their effort carrying out tasks and the overarching goal over time. Through the establishment of dedicated LLM modules (blue, yellow and green blocks) for tool selection and use, reliable and effective results can be achieved, thereby benefiting from the \textit{separation of concerns} programming concept. \cite{macconnell1993code} The modular approach afforded by our Agentic Framework permits easier development, debugging, and extension to more complex future tasks. Further, typically Agentic frameworks have more, and strategically placed, human-designed \textit{system prompts} -- shown by the red thought bubbles in Fig. \ref{fig1}. A judiciously designed, well-prompted, Agentic Framework can lead to significantly improved performance over a single LLM, thereby guiding it's behavior to engage in more exploratory actions, including autonomous novel research.

Recent works have explored the potential of LLMs to perform science tasks. For example, research Agents were shown capable of exploring research ideas, running experiments, and generating reports in an automated loop.\cite{ai_sci,agent_lib}  They focused on general scientific tasks but lacked specialized integration for metamaterial inverse design. Another approach proposed an agentic system for the inverse design of metalenses, but relied on a pre-coded forward model, and therefore lacks flexibility to carry out more general tasks.\cite{aim} In contrast, our Agentic Framework supports the full end-to-end process, from forward model training to inverse design, enabling a flexible and extensible system. The user specifies an electromagnetic property that they are interested in achieving and describes this to the Agent (Planner shown in blue) via the input (red) -- see Fig. \ref{fig1}. The Planner then carries out the process of achieving this task utilizing the specialized LLMs and tools as shown in Fig. \ref{fig1}. The end output (red) result is the description of the geometry and material properties of a metamaterial photonic system that achieves the desired goal. An established benchmark metamaterial photonic system\cite{deng2021benchmarking} is used to demonstrate the effectiveness of our framework, thereby verifying its internal reflection capabilities, versatility, and success in inverse design. The Agentic Framework presented here is not limited to the photonic metamaterial system demonstrated and can be readily extended to other material and spectroscopic systems.

\subsection{Planner}
The Planner Agent gathers essential information from the user to plan the overall research task.  During this process the Planner will ask the user for a target spectrum $s^{*}$ and task details.  Specifically, it seeks to understand the task background, primary research objectives, and dataset information. Once the Planner gathers these details, it creates a plan that divides the overall goal into tasks and specifies which agent will handle each. For example, the Planner can decide to:
\begin{itemize}
    \item Develop a forward model
    \item Perform inverse design
    \item Develop both a forward model and perform inverse design
\end{itemize}

The Planner synthesizes a high-level initial strategy which it stores in Memory. We note that this is not a fixed script -- the Planner is actively managing each step of the research process and evaluating the results. The outcome of each task is then assessed and the Planner decides the next step that it believes has the best chance for success. The iterative cycle of planning, execution, and analysis of results is what enables our Agentic Framework to adapt its course of action and which may lead to varied and potentially novel results.

The Planner Agent additionally generates a detailed task description for AI-Driven Exploration (AIDE), which is an LLM coding agent for machine learning tasks.\cite{jiang2025aide} This description includes input/output dimensions, evaluation metrics, and possible solutions to ensure controllable code generation.

\subsection{Input Verifier}
The Input Verifier agent ensures that all necessary external files and code inputs required by the process are available and valid. For example, forward modeling requires some means of generating data for model training and validation (e.g., numerical simulation); and inverse design requires a target spectrum file. The input verifier calls the $\text{file\_check}$ tool, which accepts one or more file paths as input, checks their existence, and returns whether each exists. If any of these are missing, the Input Verifier will prompt the user to provide or correct these inputs.

\subsection{Forward Modeler}
\label{Forward_R}
\begin{figure}[ptb]
    \centering
    \includegraphics[width=0.9\textwidth]{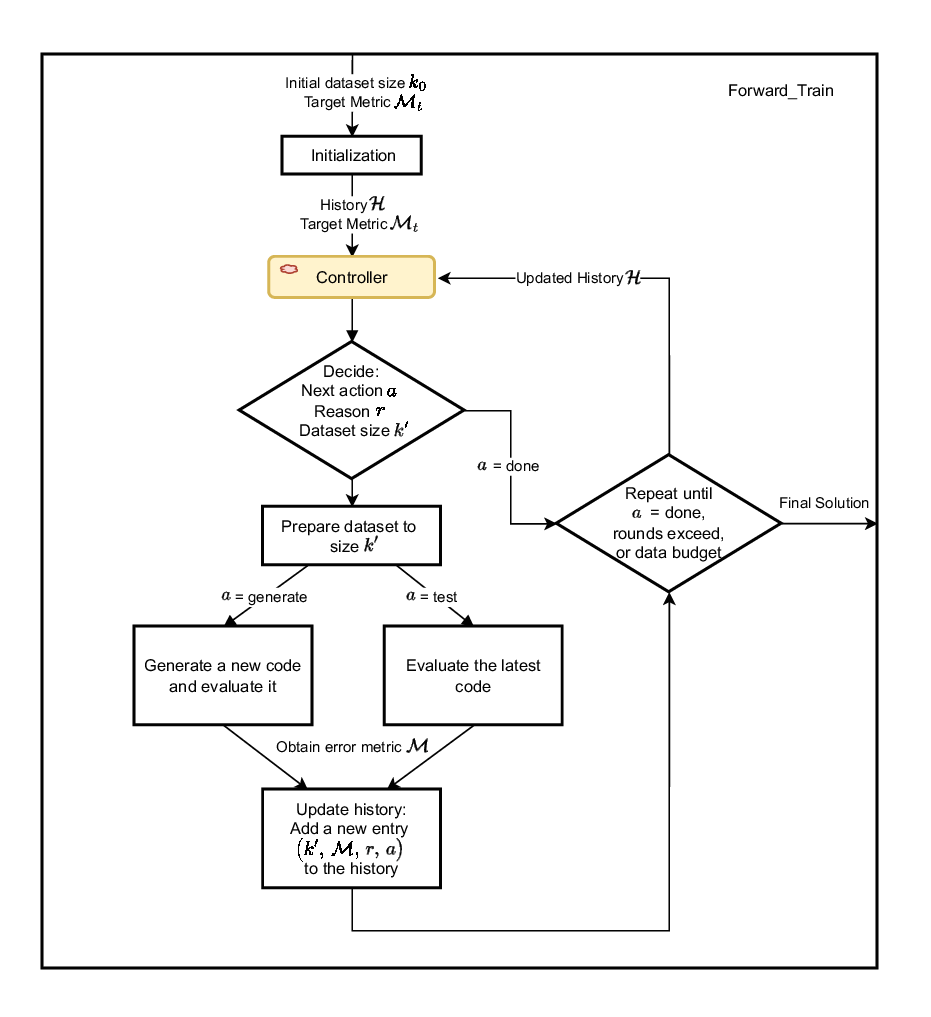}
    \caption{Flowchart of the autonomous Forward$\_$Train optimization loop. The process begins with Initialization, which builds the first training set of size \(k_0\), generates and evaluates the initial model (\(\mathcal{M}_0\)), and logs that event \((k_0, \mathcal{M}_0, \text{Initialization}, \texttt{generate})\) in the history \(\mathcal{H}\). On each iteration, the Controller reads \(\mathcal{H}\) and the target metric to choose an action—\texttt{done}, \texttt{generate}, or \texttt{test}—and a new dataset size \(k'\). After preparing the dataset, the Decision node either stops, produces new model code via AIDE (and evaluates it), or tests the current model. The outcome \((k', \mathcal{M}, r, a)\) is appended to \(\mathcal{H}\), and the cycle repeats until action is \texttt{done}, the data budget or maximum rounds is reached.
}
    \label{fig2}
\end{figure}

The Forward Modeler Agent is designed to develop a deep learning surrogate model via the tool Forward$\_$Train shown in Fig. \ref{fig1}. The internal details of the Forward$\_$Train tool are presented in Fig. \ref{fig2}. Forward$\_$Train consists of an iterative loop that intelligently balances the competing costs of data acquisition (e.g., from numerical simulation) versus model architecture optimization, which requires training and validation of candidate surrogate models. This entire process is governed by a specialized LLM called the Controller, whose logic is detailed in Algorithm \ref{alg:aide-loop}.

The loop is initialized with an initial dataset of size $k_0$, and a target performance metric $\mathcal{M}_t$. The algorithm uses these inputs to prepare an initial dataset $D_0$ with total size $k_0$, where a training:validation split of 10:1 is used.  An initial surrogate model, denoted $\texttt{code}_0$, is automatically generated using AIDE \cite{jiang2025aide} (described below), and the performance of this surrogate model is denoted $\mathcal{M}_0$. This “initialization” event \((k_0, \mathcal{M}_0, \text{Initialization}, \texttt{generate})\) is recorded in a new history json log \(\mathcal{H}\). The controller sets an input target metric $\mathcal{M}_t$ as its goal, which will then govern the main iterative loop.

At each iteration, the Controller analyzes the performance history ($\mathcal{H}$) and makes a strategic decision. This decision is a hybrid of learned reasoning and hard-coded rules. As governed by its prompt, the Controller is guided to choose progressively larger values for the size of the data set $k'$, to ensure progress. A new dataset $D'$ containing $k'$ samples will be generated (\textsc{Prepare Dataset} in the Algorithm \ref{alg:aide-loop}). The Controller also outputs an action, $a$, which can be \texttt{generate}, \texttt{test}, or \texttt{done}. The action \texttt{test} executes \textsc{Evaluate}， which runs the current latest model $\texttt{code}_{latest}$  on the new dataset and logs its performance $\mathcal{M}$ (\textsc{Evaluate} in the Algorithm \ref{alg:aide-loop}). The action \texttt{generate} invokes AIDE to produce a new model and evaluate it, returning the new code $\texttt{code}_{latest}$ and its performance $\mathcal{M}$ (\textsc{Code Generation} in the Algorithm \ref{alg:aide-loop}). The \texttt{done} action is the LLM's determination that the process has converged or that the target metric $\mathcal{M}_t$ has been reached. The loop also terminates if it hits a hard-coded limit, such as exceeding the maximum data budget or number of rounds (e.g. max rounds = 50; data budget = 50,000 samples), as specified in the algorithm's termination condition. The Controller also outputs its plain-text reasoning ($r$) which is a short text description explaining why a particular action was chosen. For each iteration, the event $\bigl(k',\,\mathcal{M},\,r,\,a\bigr)$ is appended to history $\mathcal{H}$. 

\textsc{Prepare Dataset} This tool accepts a target size $k'$ and at each iteration generates only the additional $(k' - k_{\mathrm{prev}})$ geometry–spectrum pairs required to grow the existing dataset up to $k'$, where $k_{\mathrm{prev}}$ is the size of the dataset in the previous iteration. Those data may be an existing dataset or produced by a simulation API tool. In our case, the new data are generated by a user-provided function (e.g., numerical simulation) and then they are automatically split into training and validation subsets in a $10:1$ ratio. The generated dataset $D'$ is then handed off to the next stage of the process selected by the Controller: either \texttt{test} to test the current code or \texttt{generate} to generate a new one.

\textsc{Code Generation}:
When the Controller's decision is to \texttt{generate}, it utilizes AIDE\cite{jiang2025aide} -- a state-of-the-art coding agent\cite{chan2025mlebench} -- to design a new DNN forward model. Based on a task description provided by the Planner, AIDE is permitted to autonomously investigate various model architectures, such as residual neural networks and transformers, to find the best-performing model for the data at hand. This step will also test the new solution $\mathtt{code}_{latest}$ on the latest dataset and obtain the model performance $\mathcal{M}$. 

\begin{algorithm}[t]
\caption{Agentic Loop inside Forward$\_$Train}
\label{alg:aide-loop}
\begin{algorithmic}[1]
\Require initial size $k_{0}$, target Metric~$\mathcal{M}_t$
\State \textbf{Initialize:} history $\mathcal{H}\leftarrow\varnothing$
\State 
$D_0 =\textsc{Prepare Dataset}(k_0)$
\State $(\mathcal{M}_{0},\,\mathtt{code}_{latest})\leftarrow\textsc{Code Generation}(D_{0})$ \Comment{Generate solution on initial dataset}
\State $\mathcal{H}\leftarrow\mathcal{H}\cup\bigl(k_{0},\,\mathcal{M}_{0},\,\text{Initialization},\mathtt{generate}\bigr)$

\Repeat
    \State $(a,\,k',r)\leftarrow\text{Controller}(\mathcal{H},\mathcal{M}_t)$  \Comment{Get action, new size, and reason}
    \State
    $D' =\textsc{Prepare Dataset}(k')$
    \If{$a=\mathtt{done}$}
        \State \Comment{LLM decides to terminate the loop}
        \State \Return $\mathtt{code}_{latest}$ 
    \ElsIf{$a=\mathtt{generate}$}
        \State \Comment{Generate and evaluate a new code}
        \State $(\mathcal{M},\,\mathtt{code}_{latest})\leftarrow\textsc{Code Generation}(D')$
    \ElsIf{$a=\mathtt{test}$}
        \State \Comment{Evaluate the latest code on a larger dataset}
        \State $\mathcal{M}\leftarrow\textsc{Evaluate}(\mathtt{code}_{latest},\,D')$
    \EndIf
    \State $\mathcal{H}\leftarrow\mathcal{H}\cup\bigl(k',\,\mathcal{M},\,r,\,a\bigr)$ \Comment{Update history}
\Until{budget for data or rounds is exceeded} \Comment{Hard-coded termination}
\State \Return $\mathtt{code}_{latest}$ \Comment{Return the latest model code}
\end{algorithmic}
\end{algorithm}

\subsection{Inverse Designer}
The primary goal of the Inverse Designer Agent is to solve the inverse problem: given a desired spectral response ($s^*$), it designs a metamaterial geometry ($g^*$) that reproduces the target spectrum $s^*$. While AIDE achieves accurate results in forward prediction, it is not suited for inverse design tasks. As shown in the Supporting Information \ref{supp}, a human designer can achieve an inverse result that is greater than two orders of magnitude better than that achievable with AIDE. Therefore in our study we use the NA\cite{NEURIPS2020_007ff380} as an API tool Neural$\_$Adjoint for agent use.  The NA method repurposes a forward surrogate model for effective inverse design, without any additional training.  Therefore, once the forward surrogate model is trained, it can be utilized as a static tool for inverse design. 

\begin{figure}[ptb]
    \centering
    \includegraphics[width=\textwidth]{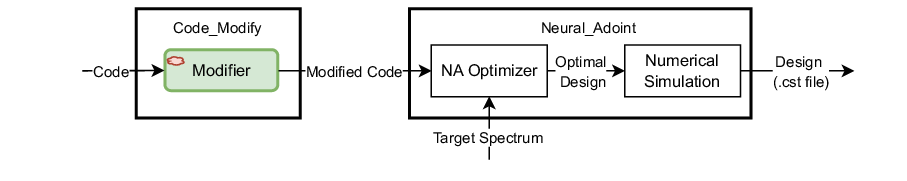}
    \caption{Flowchart of the Inverse Designer. The Inverse Designer first passes the Python code generated by the Forward Modeler to the tool Code$\_$Modify. Inside Code$\_$Modify, a LLM called Modifier is used to adapt the code for inverse design. This modified code, and the user-specified target spectrum, are fed into the Neural$\_$Adjoint tool. the NA Optimizer uses gradient descent to generate an optimal design. Finally, the optimal geometry goes to Numerical Simulation (CST Microwave Studio), which produces the corresponding .cst project file. }
    \label{fig3}
\end{figure}

After forward modeling, the Inverse Designer agent (see Fig. \ref{fig1}) will use the Code$\_$Modify, which uses an LLM ("Modifier" in Fig. \ref{fig3}) prompt to rename the model and save both the trained network parameters and data scaler  in the format expected by tool Neural$\_$Adjoint. Inside the Neural$\_$Adjoint tool, the gradient optimizer ("NA Optimizer" in Fig. \ref{fig3}) then generates a set of candidate design solutions by NA and returns the geometry with the lowest error between the forward model prediction and the target spectrum (provided by user) as the optimal design. If a Computational Electromagnetic Simulation (CEMS) tool is available, the Inverse Designer could choose to verify the optimal design via adjusting the geometry parameters on a user-supplied CST file and export the design (.cst) file ("Numerical Simulation" in Fig. \ref{fig3}).

\section{Results and Discussion}

To validate our Agentic Framework, we test its performance on the complex task of modeling an all-dielectric metamaterial (ADM) and subsequently determining an inverse design needed to achieve a desired spectra -- an established benchmark problem.\cite{deng2021benchmarking,deng2021neural}. Crucially, this benchmark problem has been previously solved by human scientists, thereby allowing us to compare the results of our agentic system to those obtained by humans.  The ADM unit-cell consists of a $2\times2$ array of elliptic resonators characterized by fourteen geometric parameters: height ($h$), periodicity ($p$), semi-major axes ($r_{ma}$), semi-minor axes ($r_{mi}$), and rotational angles ($\theta$).\cite{Deng2021} The capabilities of our agentic system are evaluated on two distinct experimental settings where the main difference is in the forward modeling process: (i) Target MSE and (ii) Fixed Large Dataset. The first experiment evaluates the autonomy and decision-making capabilities of the agentic system in the process of developing a forward surrogate model by providing a target error metric -- Mean Squared Error (MSE) -- and allowing the agent to dynamically manage model generation and optimization versus dataset size. The second experiment evaluates the capability of the agent to develop an accurate forward model given a large fixed dataset, which is a useful setting because it allows us to compare to results obtain by human experts in prior research conducted under similar settings \cite{deng2021benchmarking}.In both experimental settings, once the forward surrogate meets its respective training criterion, our agentic will perform the inverse design task.

\begin{table}[ptb]
\centering
\caption{Fixed dataset size results. Comparison between forward solutions obtained by Human designers and our Agentic Framework. Here TrL means Transformer layer, RMLP means Residual MLP, and SE means Squeeze-and-Excitation. \textdagger The average inverse MSE was not directly reported\cite{deng2021neural}, but a representative value was show in Fig. 3 of paper \cite{deng2021neural}. \textdaggerdbl The forward MSE was not directly reported in paper\cite{Ren2022}, but was found in the associated GitHub repo.
}
\newcolumntype{C}[1]{>{\centering\arraybackslash}p{#1}}
\begin{tabular}{l C{3.2cm} C{2.2cm} C{1.9cm} C{2.2cm} C{2.2cm}}
\toprule
 &  Neural Network Architecture &  Training Loss &   Param Size ($\times10^6$) & Forward MSE ($\times10^{-3}$) &  Inverse MSE ($\times10^{-3}$)\\
\midrule
Agent & TrL + RMLP & Smooth L1 & 5.3 & 1.3  &  1.8  \\
Agent & SE–RMLP & 	MSE & 1.1 & 1.5  & 1.4  \\
Human\cite{deng2021neural} & MLP+1D Conv & MSE & 11.6 &  1.2  & 0.94\textdagger\\
Human\cite{Ren2022} & RMLP & MSE & 40.1 &  1.2\textdaggerdbl & 0.3 \\
\bottomrule
\end{tabular}
\label{tab:Table1}
\end{table}

\subsection{Target MSE}
The process begins with a user query to the Planner (e.g., 'Design and optimize a deep learning regression model to predict the electromagnetic spectrum from geometry parameters using supervised learning with the MSE target $2\times10^{-3}$, then perform the inverse design on the target spectrum'). The Planner reads the prompt, asks the user to supply missing information (e.g., input/output dimensions and the path to the target spectrum file), and then verifies the required inputs by Input Verifier. Once all inputs are validated, the Planner delivers the forward modeling process to the Forward Modeler.

The forward modeling task begins with a small initial dataset — similar to what a human designer may have at the beginning of a study. As the training progresses, the dataset is autonomously expanded in order to try to meet the target MSE. In practice, generating a large training dataset through numerical simulations could take a few months. To speed up the process, each time the agent requests additional data, it will draw a subset of size k' from an existing ADM dataset.\cite{deng2021neural} This sampling method is functionally identical to running new simulations as the agent still receives new geometry–spectrum pairs when needed. In this case, our system's decision logic remains the same as the real simulation data generation, without the large computational costs of a real simulator.

\begin{figure}[ptb]
    \centering
    \includegraphics[width=0.8\textwidth]{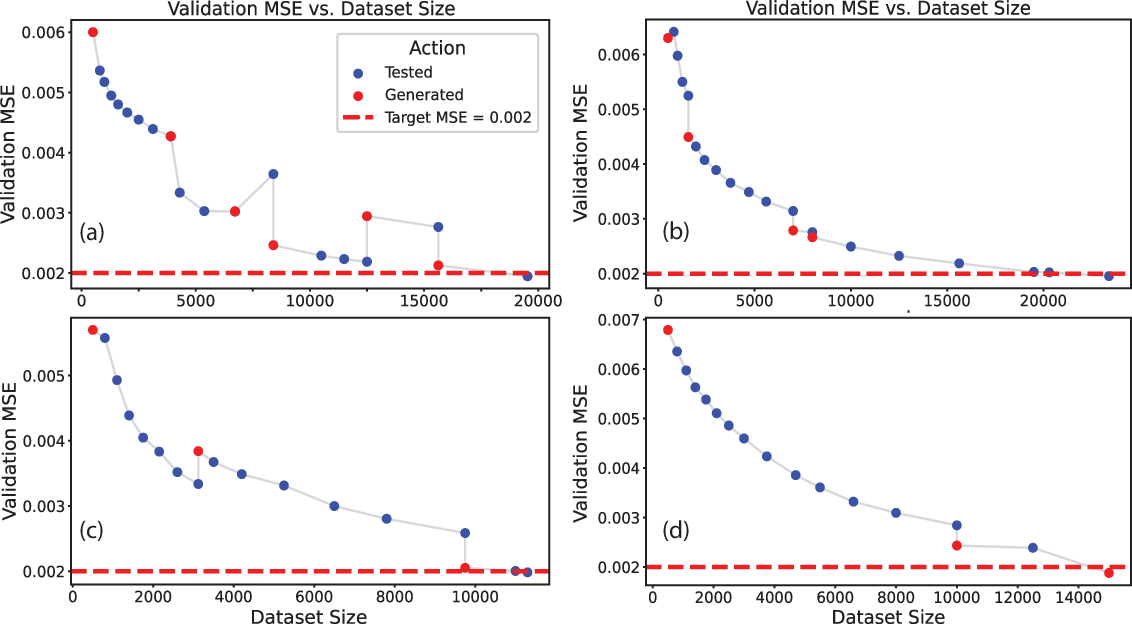}
    \caption{Dependence of validation MSE vs dataset size for four different experiments with identical initial conditions.}
    \label{fig5}
\end{figure}

Some examples of our experiments results are shown in Fig. \ref{fig5}. In Fig. \ref{fig5}(a), for example, we find that validation MSE decreases rapidly with increasing dataset size, and therefore in this instance the Agentic system has decided to continue to generate data. After reaching approximately 3,900 samples, the improvements in MSE become inefficient, so the system chooses to generate new model code (red points in Fig.\ref{fig5} (a)). The agent's reason for this action -- as internal dialogue shows -- is given as ``\textit{diminishing returns suggest saturation; regenerate code}". Moreover, when code generation gives a solution with worse performance, the system may choose to regenerate the solution. During this entire process, the Agent continuously evaluates whether it can achieve the target MSE. For example, the agent provides the following reason for retaining the current forward surrogate architecture rather than developing a new one: ``\textit{At $\displaystyle \sim6.45\times10^{-7}\,\Delta \mathrm{MSE}$ \textbackslash ${\Delta\text{sample}}$, approximately $450$--$1000$ more samples should reduce MSE to $\le0.002$; test with moderate data growth.}" In the example shown in Fig. \ref{fig5} (a) we find that the Agent achieves the target MSE with a dataset size of around 20,000. These results illustrate that the Agent is capable of autonomously achieving its goal, making systematic and progressive improvements rather than acting randomly.

To study the robustness and generalization capabilities of our Agent for forward design, we repeated this experiment multiple times. These additional runs -- see Fig.\ref{fig5} (b)(c)(d) -- all reached the forward target validation MSE of $2\times10^{-3}$. We find that the Agent achieved its goal but often used different neural network architectures, and needed varying dataset sizes in each case. Specifically the same validation MSE required dataset sizes of approximately 24,000, 11,500, and 15,000 in the experiments shown in Figs. \ref{fig5} (b), (c), and (d), respectively. These variations as well as minor differences in final MSE indicate that the system effectively makes decisions based on the specific progress made in each run, showing good robustness and generalization.

Following completion of the forward model, the Agentic system invokes the Inverse Designer, which leverages NA on the forward model to optimize geometry parameters that produce a target spectral response. The generated geometries are then evaluated using the re-simulation error between the final spectrum provided by the Agent and the ground truth from numerical simulations. To evaluate the ability of our Agentic Framework to achieve optimal inverse designs, we use a test dataset size of 100 target spectra. We find that the inverse-designed geometry parameters yield an average re-simulated MSE of $1.7\times10^{-3}$ relative to the target spectrum, and results are shown in Fig. \ref{fig6} (a)(c). We note that the result yields an accurate metamaterial geometry giving an average validation error of $\sim 0.2 \%$ in the spectrum.

\begin{figure}[ptb]
    \centering
    \includegraphics[width=0.7\textwidth]{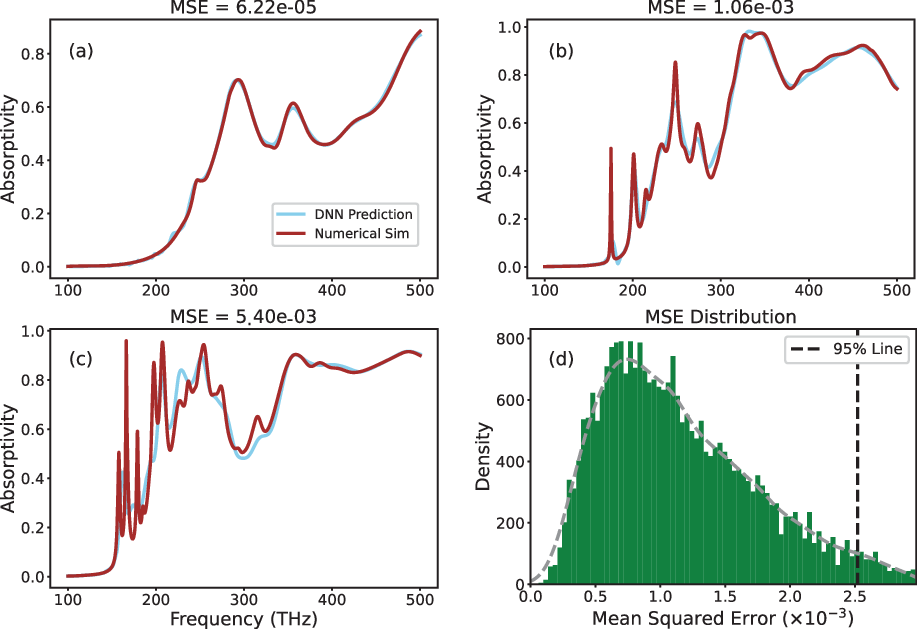}
    \caption{Forward model mean-squared-error (MSE) between prediction (blue) and ground truth numerical simulations (red). The Agentic Framework's solution result with (a) the best, (b) median, and (c) the worst. (d) MSE distribution for the test dataset.}
    \label{fig4}
\end{figure}

\subsection{Fixed Large Dataset}

To further compare our Agentic Framework against expert human-designed models, we test its performance on a large, fixed dataset of 42,250 geometry-spectrum training pairs. This experiment is meant to mirror a common scenario faced by a human designer where they use a pre-existing dataset, or they have collected as much data as they can within a fixed computational budget. Unlike the previous experiment, since the agent possesses a dataset size that is already at the specified max data budget, it therefore forgoes data expansion and instead focused only on model optimization and evaluation. All other components of the experiment remain unchanged. One model developed during these experiments consisted of a four-layer transformer encoder (TR Layer) followed by a three-layer residual multi-layer perceptron (RMLP) with one skip connection between the input to the RMLP and RMLP's final hidden layer. The MSE achieved on the unseen test dataset is $1.3\times10^{-3}$. Detailed results, including examples and the histogram of MSE across the test set, are shown in Fig.~\ref{fig5}(a)–(d). In another trial, the developed model begins with a linear layer that maps 14 input features to a hidden dimension of 256, processes them through four residual blocks which includes a squeeze-and-excitation (SE) in each block, and projects to a 2001-dimensional spectrum by a linear layer with achieved MSE $1.5\times10^{-3}$ and we summarize the results in Table \ref{tab:Table1}.

In comparison, using the same dataset a human expert designed a ten-layer fully connected network with four 1D convolution layers in one study,\cite{deng2021neural} and in another a ten-layer MLP.\cite{deng2021benchmarking} (We note that at the time those works were published, Transformers had not yet been widely-used in metamaterials research.) Notably, whereas a human designer may choose to use a neural network architecture that is proven to work and simple, the agent explored a broad design space consisting of various types of deep learning models. In both studies a value of $1.2\times10^{-3}$ was obtained in the same test set, indicating that the agent-developed model is relatively close to those obtained by expert human designers. 

\begin{figure}[ptb]
    \centering
    \includegraphics[width=0.7\textwidth]{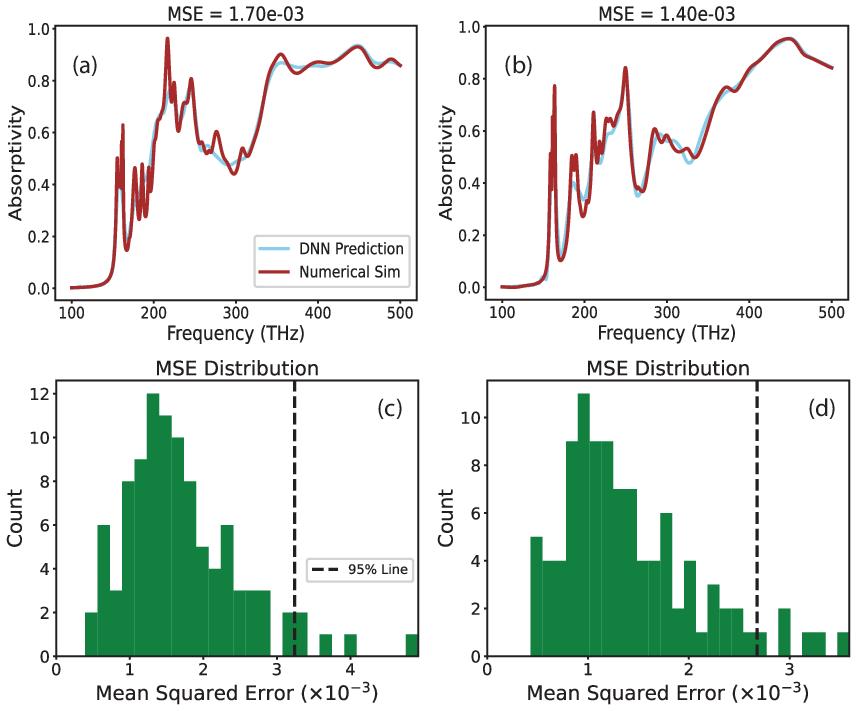}
    \caption{Inverse resimulation MSE between prediction and numerical simulations. We show a representative average result for (a) the target MSE experiment, and (b) the fixed dataset size experiment. In (c) and (d) we show the corresponding distributions of the (a) and (b) experiments, respectively. The dashed vertical line represents the delineation of 95\% of the data.}
    \label{fig6}
\end{figure}

In the subsequent inverse design step, the Inverse Designer applies the NA tool. We list the results on a test set with 100 samples in Table \ref{tab:Table1}. In Fig. \ref{fig6} we also show the more detailed inverse results for agent SE-RMLP trial: the inverse prediction (blue curve) and the ground truth numerical simulation (red curve) for a case which achieves an approximate average MSE. In Fig.\ref{fig6}(d) we show the test set distribution in Fig.\ref{fig6}(d). Compared with human-designed inverse models\cite{deng2021neural,Ren2022}, our Agentic system produces a higher inverse error. This is because we currently implement Neural$\_$Adjoint as a pre-coded tool with fixed hyperparameters (e.g., number of backpropagation steps). We believe that if we endow the agent with the ability to fine-tune the NA loop, that inverse performance could be improved.

\section{Conclusion}

In this work we proposed, developed, and demonstrated an Agentic Framework for the automated inverse design of metamaterial systems. The framework consists of a Planner agent to orchestrate the process, an Input Verifier to check the integrity of the input, a Forward Modeler to develop deep learning surrogate models, and an Inverse Designer to propose geometries that achieve the goal.

Through experiments on all-dielectric metasurfaces, we demonstrated that the framework is able to dynamically adapt its strategy at each stage and autonomously control the entire design process. Our Agentic Framework consistently achieves performance comparable to human expert-designed solutions. These results verify the automation capabilities of our Agentic Framework approach, highlighting its potential to accelerate deep learning research in metamaterials and related areas of inquiry, thereby reducing the expertise and time requirements.

Future work will focus on enabling self-improvement capabilities\cite{zhang2025darwin} i.e. endowing the Agentic Framework with the ability to create new tools and agents themselves, thereby pushing the system towards the grand goal of achieving an Agentic Researcher.






\begin{suppinfo}
\label{supp}
Each rounded rectangle shown in Fig. \ref{fig1} consists of a large language model (LLM). Each of these LLMs possess a system prompt -- as denoted by the red thought bubble -- carefully designed to enable the complex forward and inverse design process. We detail all system prompts for the Planner, Input Verifier, Forward Modeler, and Inverse Designer. 

\end{suppinfo}

\bibliography{references}

\end{document}